\documentclass{./styles/svproc}

\usepackage{graphics} 
\usepackage{epsfig} 
\usepackage{times} 
\usepackage{amsmath} 
\usepackage{amssymb}  
\usepackage{xcolor}

\usepackage{booktabs}
\usepackage{multirow}
\usepackage{cite}
\usepackage{url}

\begin{document}
\mainmatter              
\title{EL3DD: Extended Latent 3D Diffusion for Language Conditioned Multitask Manipulation}
\titlerunning{EL3DD}  
%
\author{Jonas Bode\inst{1} \and Raphael Memmesheimer\inst{1} \and
Sven Behnke\inst{1}}
\authorrunning{Bode, Memmesheimer, and Behnke} 
%
\tocauthor{Jonas Bode, Raphael Memmesheimer, Sven Behnke}
\institute{$^1$ Autonomous Intelligent Systems, University of Bonn\\
\email{\{bode, memmesheimer\}@ais.uni-bonn.de}, \email{behnke@cs.uni-bonn.de}\\
home page:
\texttt{https://www.ais.uni-bonn.de/}}

\maketitle              

\begin{abstract}
Acting in human environments is a crucial capability for general-purpose robots, necessitating a robust understanding of natural language and its application to physical tasks. This paper seeks to harness the capabilities of diffusion models within a visuomotor policy framework that merges visual and textual inputs to generate precise robotic trajectories. By employing reference demonstrations during training, the model learns to execute manipulation tasks specified through textual commands within the robot's immediate environment. The proposed research aims to extend an existing model by leveraging improved embeddings, and adapting techniques from diffusion models for image generation. We evaluate our methods on the CALVIN dataset, proving enhanced performance on various manipulation tasks and an increased long-horizon success rate when multiple tasks are executed in sequence. Our approach reinforces the usefulness of diffusion models and contributes towards general multitask manipulation.
\end{abstract}
\section{Introduction}\label{chap:introduction}

The development of general-purpose robots capable of adapting to human-centered environments has long been a goal in robotics, demanding advancements in perception, reasoning, and manipulation. Effective interaction within settings such as autonomous service or assistive robotics requires the ability to interpret natural language instructions and translate them into context-aware actions. Recent advances in the utilization of Large Language Models (LLMs) and Vision-Language Models (VLMs) to break down high-level natural language tasks into smaller steps are proving themselves to be increasingly viable~\cite{bode2024prompting, Memmesheimer:Winner2024, GPT}. However, while VLMs can assist in reasoning as well as the interpretation of natural language commands, they are insufficient to translate low-level steps such as "Pick up the apple" into actionable trajectories. 

Through imitation learning, current advances are enabling robots to perform complex tasks by observing demonstrations and responding to new instructions in a zero-shot or few-shot fashion~\cite{PerAct, HierarchicalDiffusionPolicy, 3dda, CALVIN, DiffusionPolicy}. These Vision-Language-Action Models (VLAs) have been augmented by advances in computer vision, natural language processing, large-scale diffusion models.

\begin{figure}[]
	\centering
	\includegraphics[width=0.95\linewidth]{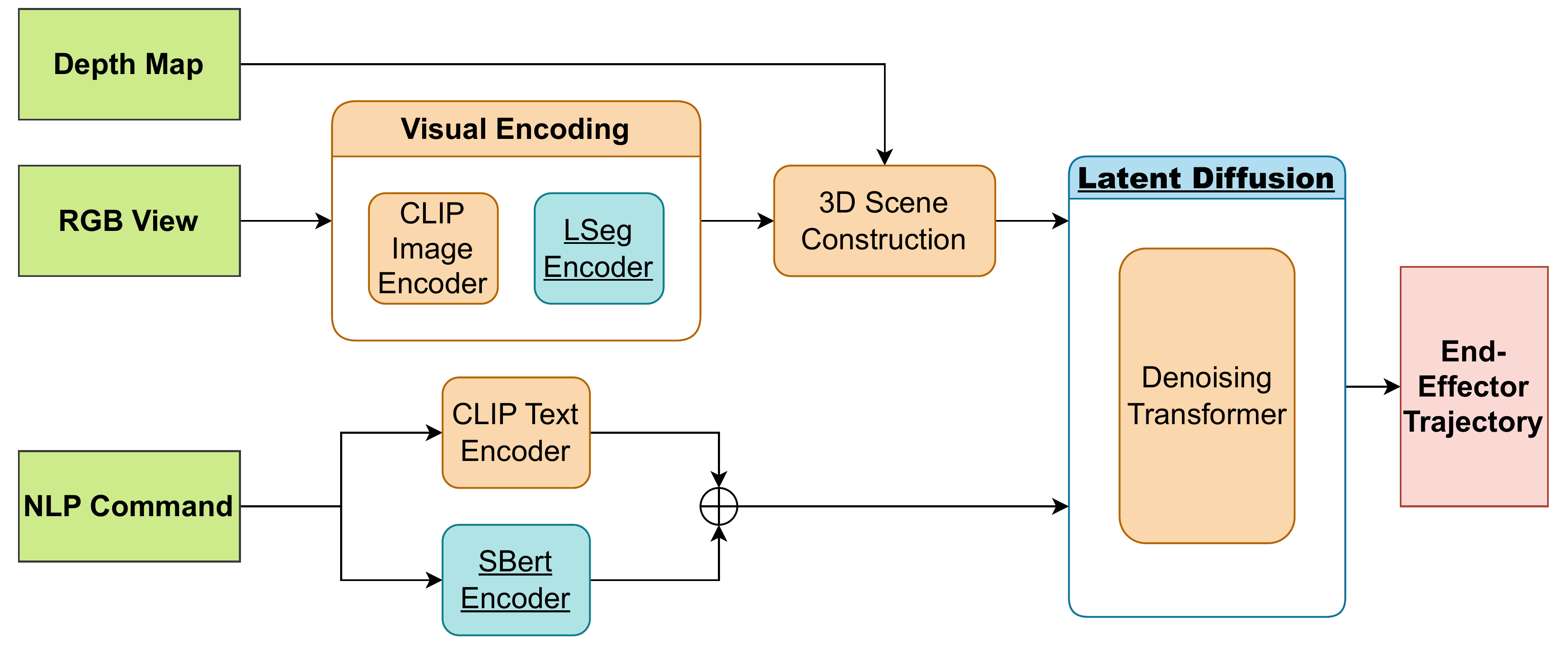}
	\caption{Overview of our EL3DD architecture. Compared to 3DDA~\cite{3dda} we replace the CLIP Image encoder with an LSeg~\cite{Lseg} image encoder, add an additional semantic S-BERT~\cite{SentenceBert} encoding, and expand the denoising transformer, which generates the end-effector trajectories into an LDM. The figure shows input data in green, output data in red, components carried over from 3DDA in orange and new components in blue.}
	\label{fig:teaser}
	\vspace{-10pt}
\end{figure}

Diffusion models have shown remarkable success in generative tasks, particularly within image generation~\cite{DenoisingDiffusion, LatentStableDiffusion}. Their robustness in representing complex data distributions makes them well-suited for multimodal fusion of visual and textual information in robotic tasks~\cite{DiffusionPolicy}. Through language-conditioned diffusion models, human commands can dynamically influence robot behavior, ensuring that manipulation actions are contextually and semantically grounded.

This paper aims to improve upon one existing VLA, the 3D Diffuser Actor (3DDA)~\cite{3dda}. The model combines natural language instructions and a 3D scene representation to ground the generation of end-effector trajectories. We advance 3DDA in this work through additions regarding semantic embeddings, the perception backbone, and modification of the diffusion model. Figure~\ref{fig:teaser} shows an overview of the proposed architecture.

To evaluate the proposed framework, we use the CALVIN dataset~\cite{CALVIN}, which provides a comprehensive set of long-horizon manipulation tasks in diverse environments and zero-shot evaluation scenarios in novel scenes.

Our contributions in this paper can be summarized as:

\begin{enumerate}
	\item We propose incorporating an additional S-BERT~\cite{SentenceBert} text encoder into the 3DDA architecture to enhance language representation.
	
	\item We replace the vision backbone with LSeg~\cite{Lseg}, integrating per-pixel CLIP embeddings for improved scene understanding.
	
	\item We redesign the diffusion component of 3DDA to leverage an LDM.
	
	\item We demonstrate the Extended Latent 3D Diffuser architecture (EL3DD), an incremental improvement over 3DDA, through evaluation on the CALVIN dataset~\cite{CALVIN}. 
\end{enumerate}

We begin with an overview of related works in Section~\ref{chap:related}. Next, we detail the 3DDA model~\cite{3dda} and our proposed improvements in Section~\ref{chap:methodology}. Section~\ref{chap:evaluation} presents our experiments and results, followed by a summary of our contributions in Section~\ref{chap:conclusion}.

\section{Related Works}\label{chap:related}

Recent advancements in LLM capabilities have spurred growing research into general-purpose robotics. Many approaches leverage LLMs to generate pseudocode or structured function calls based on user input, which can then be parsed by algorithms and executed on a robot~\cite{bode2024prompting, Memmesheimer:Winner2024}. While promising, this approach constrains LLMs to a predefined syntax for interacting with the world. This constraint is eliminated by VLAs which directly output actionable end-effector or joint trajectories.

This section reviews existing research on language-conditioned multitask manipulation and robotic manipulation using diffusion models

\subsection{Language-Conditioned Multitask Manipulation}

Conditioning robotic manipulation on natural language commands poses significant challenges. Shridhar et al. introduced a transformer-based model, PerAct~\cite{PerAct}, which employs a behavioral-cloning agent to generate end-effector trajectories. The model operates in a latent voxel space, selecting the next best voxel to move to---based on the current end-effector position. A natural language command is embedded and fed into the transformer alongside the encoded voxel data. By effectively handling high-dimensional inputs, PerAct demonstrates strong performance on the RLBench benchmark~\cite{RLBench}.

Grotz et al.~\cite{PerAct2} extended PerAct and the RLBench benchmark to address bimanual manipulation. Their architecture, PerAct$^2$, is validated in both simulation and real-world settings. While it surpasses other architectures adapted for bimanual tasks, it struggles to achieve success rates above $50\%$ for many tasks, underscoring ongoing challenges in this domain.

An alternative approach for single-arm manipulation is presented by Li et al. in GR-MG~\cite{GRMG}. This model integrates video prediction with goal image conditioning. Goal images are synthesized by modifying the current scene view using a diffusion-based image model guided by textual commands. By enabling the use of both goal images and language instructions, GR-MG expands training data to include partially annotated datasets, and pretrains on a video prediction objective.

\subsection{Manipulation Using Diffusion Models}

Since Ho et al.~\cite{DenoisingDiffusion} demonstrated the efficacy of diffusion models for high-quality image generation, these models have become a focus of extensive research. Diffusion models exhibit several desirable properties, including stability during training, robust coverage of the data distribution, generalizability, and scalability. These advantages are achieved by training a neural network to predict artificially generated noise during training. This noise prediction forms the basis for inference, where a sample of pure noise is iteratively denoised to generate outputs.

The capabilities of diffusion models have been further enhanced through innovations like latent diffusion~\cite{LatentStableDiffusion}. In this approach, the diffusion process is performed in a latent space rather than directly in the image pixel space. This latent space is accessed via a pretrained autoencoder, which reduces computational demands while improving flexibility and maintaining high fidelity.

Recently, diffusion models have also demonstrated their utility in robotic manipulation. Chi et al.~\cite{DiffusionPolicy} introduced Diffusion Policy, which employs a Denoising Diffusion Implicit Model (DDIM)~\cite{DDIP} to generate end-effector trajectories through a behavioral cloning framework. The model frames the denoising process as learning the gradient of the action-distribution under a score function derived from the data. This iterative approach enables the policy to identify appropriate action trajectories. Key innovations include the incorporation of receding horizon control, visual conditioning, and a time-series diffusion transformer, allowing real-time action inference for complex, multi-step tasks. However, this Diffusion Policy lacks language conditioning and requires retraining for each new task.

Ke et al. introduced 3DDA~\cite{3dda}, a diffusion policy for generating end-effector trajectories conditioned on 3D scene representations. By incorporating embedded language and 3D semantic information, 3DDA supports long-horizon tasks and demonstrates robust generalization across varying camera views. The model achieves great performance on both RLBench~\cite{RLBench} and CALVIN~\cite{CALVIN} for diffusion models. Given its alignment with the objectives of this paper 3DDA will serve as the baseline for further improvements proposed in this work.

\section{Methodology}  \label{chap:methodology}

We will first present an overview of the 3DDA architecture proposed by Ke et al.~\cite{3dda}, on which our approach is based on. 
Subsequently, we detail the three improvements to 3DDA considered in this paper.

\subsection{3DDA}

Given demonstration trajectories $\{(o_1, a_1), (o_2, a_2), \dots\}$ in addition to the associated task language instruction $l$, 3DDA by Ke et al.~\cite{3dda} imitates behavior through the use of a diffusion model to synthesize actions from observations. Here, $o_t$ represents the visual observations of posed RGB-D images at timestep $t$ and $a_t$ the end-effector pose at the same timestep, which can be decomposed as: $a_t = {a_t^{\text{loc}} \in \mathbb{R}^3, a_t^{\text{rot}} \in \mathbb{R}^6, a_t^{\text{open}} \in {0, 1}},$ where $a_t^{\text{loc}}$ specifies the 3D location, $a_t^{\text{rot}}$ represents the 6D rotation of the end-effector, and $a_t^{\text{open}}$ indicates the gripper state (open or closed). For better continuity. 
 
3DDA does not only predict the action of the current timestep but an entire trajectory for a temporal horizon $T$ which is denoted as $\tau_t = (a_{t:t+T}^{\text{loc}}, a_{t:t+T}^{\text{rot}})$ in addition to a series of binary open/close states $a_{t:t+T}^{\text{open}}$.

To achieve this prediction, 3DDA first constructs a 3D scene representation from the current observation $o_t$ before synthesizing the trajectories $\tau_t$ and $a_{t:t+T}^{\text{open}}$ using a denoising transformer diffusion model. 3DDA aggregates multiple RGB-D views into a single, unified 3D representation by encoding each image with a CLIP image encoder~\cite{CLIP}. The patch-wise encodings are attached to each pixel and projected into 3D space using depth information.. To incorporate task instructions, the aggregated 3D tokens undergo relative cross attention with the language task embedding, encoded using a CLIP language encoder~\cite{CLIP} and are afterwards subsampled with Farthest Point Sampling (FPS). 

The encoded input is used as input for the denoising transformer. Let $\tau^0$ denote the ground-truth trajectory, and $\tau^i$ represent the trajectory at diffusion step $i$. The noisy trajectory is formulated as:

$$\tau^i = \sqrt{\Bar{\alpha}^i} \tau^0 + \sqrt{1-\Bar{\alpha}^i} \epsilon.$$

Here, $\alpha^i = 1 - \beta^i$, $\Bar{\alpha}^i = \prod_{j=1}^i \alpha^j$, and $\{\beta^i \in (0, 1)\}^N_{i=1}$ is the variance schedule for $N$ total diffusion steps. Further $\epsilon \sim \mathcal{N}(0, 1)$ is a sample from a Gaussian distribution. The denoising transformer can now be written as $\epsilon_\theta$ making the output 

$$\hat{\epsilon} = \epsilon_\theta(\tau_t^i, i, o_t, l, c_t)$$

the estimated noise or the estimated learned gradient of the denoising process. $\tau_t^i$ is the trajectory for timestep $t$ at diffusion step $i$ and $c_t$ is the proprioception of the end-effector. If we generate a random sample $\tau_t^N \sim \mathcal{N}(0, 1)$ we can now draw a sample from our learned distribution for timestep $t$ iteratively denoising the trajectory over $N$ steps

$$\tau_t^{i-1} = \frac{1}{\sqrt{\alpha^i}} \left( \tau_t^i - \frac{\beta^i}{\sqrt{1 - \Bar{\alpha}^i}} \epsilon_\theta(o_t, l, c_t, \tau_t^i, i)\right) + \frac{1 - \Bar{\alpha}^{i+1}}{1 - \Bar{\alpha}^i} \beta^i z,$$

with $z \sim \mathcal{N}(0, 1)$ of fitting dimension.

To learn from a demonstration, a random time step $t$ and diffusion step $i$ is sampled. Then, noise is added to the ground-truth trajectory $\tau_t^i$. 3DDA generates the noise $\epsilon = (\epsilon^{\text{loc}}, \epsilon^{\text{rot}})$ for the location and rotation component of the trajectory separately. The network is then trained according to the objective

\begin{align*}
	\mathcal{L}_\theta &= w_1 \|\epsilon^\text{loc}_\theta(o_t, l, c_t, \tau_t^i, i) - \epsilon^\text{loc} \| \\
	&+ w_2 \|\epsilon^\text{rot}_\theta(o_t, l, c_t, \tau_t^i, i) - \epsilon^\text{rot} \| \\
	&+ \text{BCE} (f_\theta^\text{open}(o_t, l, c_t, \tau_t^i, i), a_{t:t+T}^\text{open}). 
\end{align*}

$w_1$ and $w_2$ are hyperparameters, while binary cross-entropy loss is used to evaluate the estimation of the binary end-effector state  $f_\theta^\text{open}$. 

The 3DDA architecture uses self and cross attention layers, modulated using FiLM~\cite{FILM}.

\subsection{Improvements}

Building on 3DDA~\cite{3dda} this paper suggests three different improvements. This section will explain all three, provide the reasoning behind the improvement and implementation details.

\subsubsection{Additional Semantic Information}

3DDA makes use of CLIP~\cite{CLIP} for both language encoding of the task instruction and the 2D RGB feature extraction. CLIP (Contrastive Language-Image Pre-training) is trained to associate text data and image contents. While this successfully encodes and reflects visual information, it might fail to sufficiently encode semantic action within the task instructions. If the task instruction is: "Push the blue block to the left," the CLIP language encoding captures the importance of the blue block and how the block has to be touched, but, being trained on still 2D images, it might not fully encode the dynamic act of pushing. This limitation is particularly relevant because CLIP encoders~\cite{CLIP} are not trained with robotics tasks in mind and are better suited for encoding still images. This is also supported by other robotics focused works like~\cite{clipfields} utilizing both CLIP and an additional text encoding.

To address this limitation, we integrate an S-BERT~\cite{SentenceBert} text encoder into the architecture. Specifically, we use "all-mpnet-base-v2" which results from a combination between~\cite{SentenceBert} and~\cite{Mpnet}. We encode the language instruction $l$ using both the CLIP text encoder and the Sentence-Bert encoder and concatenate both encodings. Through this, we can use the concatenated encodings as before without further architectural changes and apply the relative cross attention layer including the 3D tokens. 

\subsubsection{LSeg for Visual Embeddings}

3D feature tokens are created in 3DDA based on the CLIP image encoder~\cite{CLIP}, which extracts features for image patches. This can result in coarse features that lack the detail required for accurate and precise scene understanding.

We use LSeg~\cite{Lseg}, to improve upon this. LSeg is trained to generate pixel-wise CLIP embeddings for the entire image, enabling detailed and fine-grained feature extraction. For this, we exchanged the CLIP image encoder used in the original 3DDA architecture with the LSeg image encoder. 

\subsubsection{Latent Diffusion}

Latent Diffusion Models (LDMs), first proposed in~\cite{LatentStableDiffusion}, perform the diffusion process in a latent space instead of the target space. This approach leverages a pretrained encoder during training and a pretrained decoder during inference. By diffusing within the latent space, LDMs can achieve greater computational efficiency and can better capture meaningful high-level semantic content rather than more minute noise.

While 3DDA does encode the trajectories into latent space before passing them to the deeper network layers, it does not do the diffusion process within latent space. Instead, it returns to the end-effector space for each diffusion step. To integrate latent diffusion into 3DDA, we first train a joint Variational Autoencoder (VAE)~\cite{VAE} to encode and decode end-effector trajectories $\tau_t$ and end-effector state $a_{t:t+T}^\text{open}$. Next, we encode the ground-truth trajectories using the trained encoder $\mathcal{E}$ during training:

$$h_t = \mathcal{E}(\tau_t, a_{t:t+T}^\text{open})$$.

Since $h_t$ is already in latent space and combines the entire trajectory, we can get rid of the initial MLPs and linear layers used in the 3DDA architecture. 

With $h_t^i = \sqrt{\Bar{\alpha}^i} \mathcal{E}(\tau^0, a_{t:t+T}^\text{open}) + \sqrt{1-\Bar{\alpha}^i} \epsilon$, this simplifies our objective function to

\begin{align*}
	\mathcal{L}_\theta^\text{LDM} &= \|\epsilon^\text{latent}_\theta(o_t, l, c_t, h_t^i, i) - \epsilon^\text{latent} \|. \\
\end{align*}

Similarly, during inference, the LDM begins in latent space with a random sample $h_t^N \sim \mathcal{N}(0, 1)$ and is iteratively denoised over N steps

$$h_t^{i-1} = \frac{1}{\sqrt{\alpha^i}} \left( h_t^i - \frac{\beta^i}{\sqrt{1 - \Bar{\alpha}^i}} \epsilon_\theta(o_t, l, c_t, h_t^i, i)\right) + \frac{1 - \Bar{\alpha}^{i+1}}{1 - \Bar{\alpha}^i} \beta^i z.$$

The trained decoder $\mathcal{D}$ is then used to generate the wanted trajectory $(\tau_t, a_{t:t+T}^\text{open}) = \mathcal{D}(h_t^0)$.

Our modification ensures that both the diffusion process during training and the denoising process during inference occur entirely in latent space. The encoder $\mathcal{E}$ is applied at the start of each training step to encode the ground-truth trajectory into latent space. During inference, the process begins with a random sample drawn from the latent space, which is iteratively denoised. Finally, the decoder $\mathcal{D}$ converts the denoised output back into end-effector space.

\section{Experiments \& Results}\label{chap:evaluation}

In this section, we present the experimental setup, followed by a discussion of the CALVIN dataset~\cite{CALVIN}. Finally, we present the results of our experiments and discuss the implications of our findings.
To evaluate the proposed improvements, we conducted an ablation study, comparing the results of our methods to the 3DDA~\cite{3dda} baseline. We conducted our experiments using the CALVIN ABC$\to$D dataset~\cite{CALVIN}. 

We trained the models on the Marvin HPC Cluster at the University of Bonn. Each training run utilized a single SGPU node equipped with four Nvidia A100 80GB GPUs. The same hardware configuration was employed for testing and inference.

\subsection{Dataset}

The CALVIN dataset~\cite{CALVIN} is an open-source benchmark designed to support the development of language-conditioned control policies for robots. It focuses on long-horizon manipulation tasks and offers

four environments, each including various interactive elements, such as a sliding door, a drawer, buttons, switches, and differently colored blocks. This dataset encompasses $\sim 2.4$ million interaction steps and 389 unique natural language instructions over 34 manipulation tasks, providing a robust foundation for training models that generalize to complex behaviors.
We use the ABC$\to$D split for training and evaluation in this paper. It consists of training data for 3 scenes and testing data for 1 novel scene supporting zero-shot policy evaluation.

CALVIN supports RGB-D images from fixed and gripper-mounted cameras, proprioceptive data, and tactile feedback. The action space encompasses continuous control of the robot’s end-effector in Cartesian or joint space. While neither our work nor 3DDA~\cite{3dda} utilizes tactile feedback, we leverage fixed and gripper-mounted RGB-D cameras, proprioceptive data, and camera parameters.

\subsection{Ablation Study}

In Table~\ref{tab:calvin_evaluation} we compared different combinations of our three proposed improvements against the 3DDA baseline. As CALVIN~\cite{CALVIN} supports long-horizon evaluation, the table reports the success percentage for completing up to 5 tasks consecutively. This approach allows us to evaluate both the immediate likelihood of accomplishing a single task and the model's stability as it interacts repeatedly within the same environment. Additionally, the table includes the average number of completed tasks. The evaluation was done on $1000$ task chains for each model. 

Introducing S-BERT text encoding alone yields only marginal improvements over the baseline model. The small improvements that are present over the baseline are more pronounced in longer task chains. Replacing the vision backbone with LSeg significantly improves performance. This modification alone is enough to increase the average length by $0.16$ compared to the baseline and by $0.13$ compared to the S-Bert version. The combination of LSeg and S-BERT further enhances performance, achieving a $95\%$ success rate for the first task and increasing the probability of completing all five tasks by $6.2\%$ compared to the baseline.

\begin{table}[t]
	\centering
	\caption{Ablation Study for Zero-shot long-horizon evaluation on CALVIN ABC$\to$D. The baseline corresponds to the 3DDA model from~\cite{3dda}. Success percentages represent the likelihood of completing consecutive tasks in a chain.}
	\label{tab:calvin_evaluation}
	\begin{tabular}{lcccccc}
		\toprule
		\multirow{2}{*}{Applied methods}& \multicolumn{5}{c}{Tasks completed in a row} \\
		\cmidrule(lr){2-6}
		& 1 & 2 & 3 & 4 & 5 & Avg. Len \\
		\midrule
		Baseline            & 93.7 & 80.1 & 66.1 & 53.2 & 41.0 & 3.34 \\
		S-BERT              & 94.0 & 80.6 & 66.5 & 53.9 & 42.1 & 3.37 \\
		LSeg                & 94.4 & 81.5 & 68.3 & 58.6 & 47.2 & 3.50 \\
		S-BERT + LSeg       & 95.0 & 82.4 & 69.7 & 60.6 & 49.9 & 3.58 \\
		LDM                 & 95.9 & 88.3 & 77.3 & 69.8 & 58.5 & 3.90 \\
		LDM + S-BERT + LSeg & \textbf{96.6} & \textbf{90.0} & \textbf{82.4} & \textbf{74.2} & \textbf{66.2} & \textbf{4.09} \\
		\bottomrule
	\end{tabular} \\
	\vspace{-10pt}

\end{table}

The LDM versions of 3DDA perform the best out of the tested models. Using only latent diffusion lets the model outperform the best prior version of S-BERT + LSeg in all metrics. The advantage of latent diffusion becomes even clearer when observing a model with all three improvements. LDM + S-BERT + LSeg achieves the best performance by a wide margin, especially for longer task chains. It is also the only one of our models to achieve an average length above $4$ tasks.

\begin{figure}[th]
	\centering
	\includegraphics[width=0.98\linewidth]{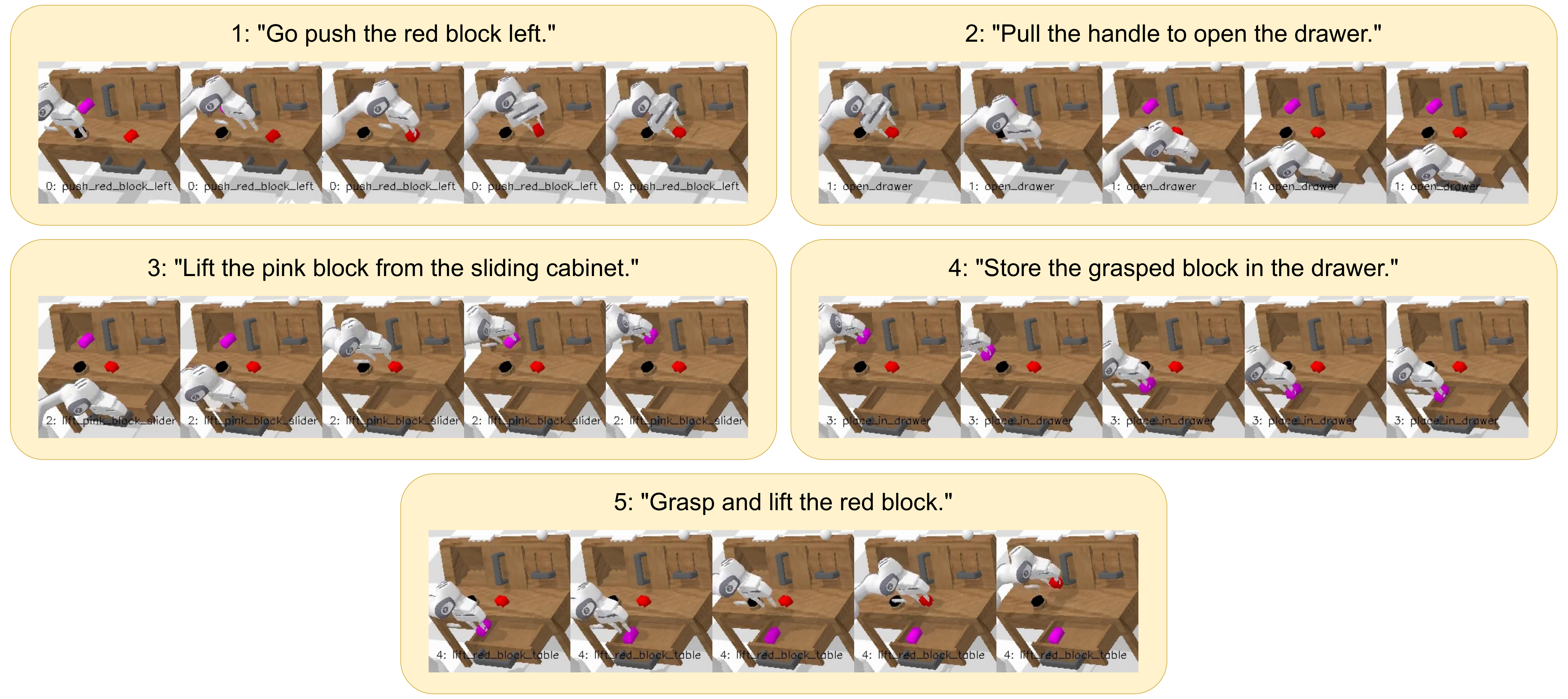}
	\caption{EL3DD executing a task chain during evaluation. All 5 tasks were successfully executed in a row.}\label{fig:example}
	\vspace{-10pt}
\end{figure}

An example of a successful five-task execution chain is depicted in Figure~\ref{fig:example}. The EL3DD model interacts with a desk in various ways, including pushing, pulling, lifting, placing, and grasping the appropriate objects.

\section{Conclusion}\label{chap:conclusion}

This paper explored language-conditioned multitask models for robot manipulation using diffusion. We proposed several enhancements to the 3DDA architecture~\cite{3dda}, contributing to the development of long-horizon VLAs capable of executing multiple tasks in sequence. Our proposed model EL3DD was evaluated on the CALVIN dataset~\cite{CALVIN} and demonstrated significant improvements over the 3DDA baseline. 

The proposed modifications enhanced the model’s language capabilities through additional S-BERT~\cite{SentenceBert} embeddings, improved its visual perception using an LSeg~\cite{Lseg} backbone to generate per-pixel CLIP embeddings, and introduced latent diffusion~\cite{LatentStableDiffusion} into the 3DDA architecture. Combining these enhancements resulted in a model that achieved great performance on the CALVIN benchmark for multitask manipulation and is the new state of the art for models directly generating end-effector actions using diffusion.

Since our investigation focused exclusively on the CALVIN dataset, it remains uncertain how well the model generalizes to other datasets with different tasks. Similarly, we did not conduct real-world tests to evaluate the trained model’s performance in less controlled environments, presenting a severe limitation to this study.

Building on the findings of this paper, future work should aim to diversify the datasets used for training and evaluation. Additionally, real-world deployment could offer valuable insights into the approach’s stability and robustness. Further, replacing the positional with a spatial encoding could improve the model’s scene understanding.

The EL3DD model proposed in this paper underscores the potential for diffusion in VLAs. While this work is limited in scope, the potential of diffusion models has yet to be pushed to its limit, offering considerable opportunities for future research.

\vspace{-5pt}

\subsubsection*{Acknowledgments}

This work has been partially funded by the Federal Ministry of Research, Technology and Space of Germany (BMFTR) in the projects "Learn2Grasp," grant 16IS21080, and "Robotics Institute Germany," grant 16ME0999.
%

\bibliographystyle{./styles/bibtex/spmpsci_unsrt.bst}
\bibliography{references.bib}

\end{document}